\def\year{2016}\relax
\documentclass[letterpaper]{article}
\usepackage{aaai16}
\usepackage{times}
\usepackage{helvet}
\usepackage{courier}

\usepackage{graphicx}
\usepackage{amsmath,bm}
\usepackage{amsfonts}
\usepackage{mathrsfs}
\usepackage{amsbsy}
\usepackage{paralist}

\DeclareMathOperator{\vect}{vec}
\DeclareMathOperator{\softmax}{softmax}

\begin{document}
\title{Convolutional Neural Networks over Tree Structures\\
for Programming Language Processing}
\author{Lili Mou,$^1$ Ge Li,$^{1*}$ Lu Zhang,$^1$ Tao Wang,$^2$ Zhi Jin$^{1*}$\\
$^1$Software Institute, Peking University \ \ \ \ $^*$Corresponding authors\\
{doublepower.mou@gmail.com}, {\{lige,zhanglu,zhijin\}@sei.pku.edu.cn}\\
$^2$Stanford University, { twangcat@stanford.edu}
}

\maketitle
\begin{abstract}
\begin{quote}
Programming language processing (similar to natural language processing)
is a hot research topic in the field of software engineering;
it has also aroused growing interest in the artificial intelligence community.
However, different from a natural language sentence, a program
contains rich, explicit, and complicated structural information.
Hence, traditional NLP models may be inappropriate for programs.
In this paper, we propose a novel tree-based convolutional neural
network (TBCNN) for programming language processing, in which a convolution
kernel is designed over programs' abstract syntax trees to capture structural
information. TBCNN is a generic architecture for
programming language processing; our experiments show
its effectiveness in two different program analysis tasks:
classifying programs according to functionality, and detecting code snippets
of certain patterns. TBCNN outperforms baseline methods,
including several neural models for NLP.
\end{quote}
\end{abstract}

\section{Introduction}
Researchers from various communities are showing growing interest
in applying artificial intelligence (AI) techniques to solve software
engineering (SE) problems \cite{AI1,development,zhanglu}. In the area of SE,
analyzing program source code---called \textit{programming language processing}
in this paper---is of particular importance.

Even though computers can run programs, they do not truly ``understand'' programs.
Analyzing source code provides a way of estimating programs' behavior, functionality,
complexity, etc. For instance, automatically detecting source code snippets of
certain patterns help programmers to discover buggy or inefficient algorithms
so as to improve code quality. Another example is managing large software
repositories, where automatic source code classification and tagging are
crucial to software reuse. Programming language processing, in fact, serves as a
foundation for many SE tasks, e.g., requirement analysis \cite{requirement},
software development and maintenance \cite{development}.

\citeauthor{naturalness} (\citeyear{naturalness}) demonstrate that programming
languages, similar to natural languages, also contain abundant statistical properties,
which are important for program analysis. These properties are difficult to capture
by humans, but justify learning-based approaches for programming language processing.
However, existing machine learning program analysis depends largely on feature engineering,
which is labor-intensive and \textit{ad hoc} to a specific task, e.g., code clone
detection \cite{sim}, and bug detection \cite{bug}. Further, evidence in the machine
learning literature suggests that human-engineered features may fail to capture the
nature of data, so they may be even worse than automatically learned ones.

The deep neural network, also known as \textit{deep learning}, is a highly automated
learning machine. By exploring multiple layers of non-linear transformation, the deep
architecture can automatically learn complicated underlying features, which are
crucial to the task of interest. Over the past few years, deep learning has made
significant breakthroughs in various fields, such as speech recognition \cite{speech},
computer vision \cite{imagenet}, and natural language processing \cite{unified}.

Despite some similarities between natural languages and programming languages,
there are also obvious differences \cite{PLNL}. 
%We argue that new architectures
%should be proposed for program analysis.  
Based on a formal language, programs
contain rich and explicit structural information. Even though structures also exist in
natural languages, they are not as stringent as in programs. \citeauthor{instinct}
(\citeyear{instinct}) illustrates an interesting example, ``The dog the stick the fire
burned beat bit the cat.'' This sentence complies with all grammar rules, but too many
attributive clauses are nested. Hence, it can hardly be understood by people due to the
limitation of human intuition capacity. On the contrary, three nested loops are
common in programs. The parse tree of a program, in fact, is typically much larger
than that of a natural language sentence---there are approximately 190 nodes on average in our
experiment, whereas a sentence comprises only 20 words in a sentiment analysis dataset \cite{RNN}.
Further, the grammar rules ``alias'' neighboring relationships among program components.
The statements inside and outside a loop, for example, do not form one semantic group,
and thus are not semantically neighboring. On the above basis, we think more effective
neural models are in need to capture structural information in programs.

In this paper, we propose a novel \textit{Tree-Based Convolutional Neural Network} (TBCNN)
based on  programs' abstract syntax trees (ASTs). We also introduce the notion of
``continuous binary trees'' and apply dynamic pooling to cope with ASTs
of different sizes and shapes. The TBCNN model is a generic architecture,
and is applied to two SE tasks in our experiments---classifying programs by
functionalities and detecting code snippets of certain patterns.
It outperforms baseline methods in both tasks, including the recursive neural
network \cite{RAE} proposed for NLP. To the best of our knowledge, this paper
is also the first to apply deep neural networks to the field of programming language 
processing.\footnote{We make our source code and the collected dataset available
through our website
(https://sites.google.com/site/treebasedcnn/).}

\section{Related Work}

Deep neural networks have made significant breakthroughs in many fields.
Stacked restricted Boltzmann machines and autoencoders are successful pretraining methods \cite{fast,laywise}. They explore the underlying
features of data in an unsupervised manner, and give
a more meaningful initialization of weights for later supervised learning with
deep neural networks.
These approaches work well with generic data (e.g. data located in a manifold embedded in a certain dimensional space), but they may not be suitable for
programming language processing, because programs contain rich structural information.
Further, AST structures also vary largely among different data samples (programs), and hence they cannot be fed directly to a fixed-size network.

To capture explicit structures in data,
it may be important and beneficial to integrate human priors to the networks \cite{RL}.
One example is convolutional neural networks
(CNNs, \citeauthor{lenet} \citeyear{lenet}; \citeauthor{imagenet} \citeyear{imagenet}),
which specify spatial neighboring information in data.
CNNs work with signals of a certain dimension (e.g., images);
they also fail to capture tree-structural information as in programs.

\citeauthor{RNN} (\citeyear{RNN}, \citeyear{RAE}) propose a recursive neural network (RNN) for NLP.
Although structural information may be coded to some extent in RNNs, the major drawback is that
only the root features are used for supervised learning, which buries illuminating
information under a complicated neural architecture.
RNNs also suffer from the difficulty of training due to the long dependency path
during back-propagation \cite{rnndifficult}.

\textbf{Subsequent work.}
After the preliminary version of this paper was preprinted on arXiv,\footnote{
On 18 September 2014 (http://arxiv.org/abs/1409.5718v1).} \citeauthor{execute} (\citeyear{execute})
use recurrent neural networks to estimate the output of restricted python programs.
\citeauthor{hoare} (\citeyear{hoare}) build recursive networks on Hoare triples.
Regarding the proposed TBCNN, we extend it to process syntactic parse trees of natural
languages \cite{tbcnn_sent}; \citeauthor{gbcnn} (\citeyear{gbcnn}) apply a similar convolutional
network over graphs to analyze molecules.

\section{Tree-Based Convolutional Neural Network}
\begin{figure*}[!t]
\centering
\includegraphics[height=.16\textwidth]{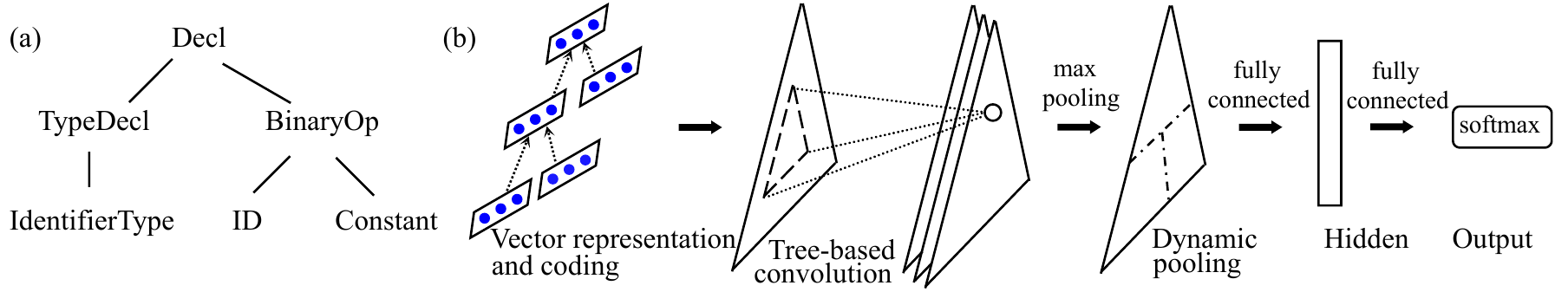}
\caption{(a) Illustration of an AST, corresponding to the {\tt C} code snippet
``int a=b+3;'' It should be notice that our model takes as input the entire AST
of a program, which is typically much larger. (b) The architecture of the Tree-Based
Convolutional Neural Network (TBCNN). The main components in our model include vector
representation and coding, tree-based convolution and dynamic pooling;
then a fully-connected hidden layer and an output layer ($\softmax$) are added.
}\label{fOverview}
\end{figure*}

Programming languages have a natural tree representation---the abstract syntax tree (AST).
Figure~\ref{fOverview}a shows the AST of the code snippet ``{\tt int a=b+3;}''.\footnote{
Parsed by pycparser (https://pypi.python.org/pypi/pycparser/).
}
Each node in the AST is an abstract component in program source code.
A node $p$ with children $c_1, \cdots, c_n$ represents
the constructing process of the component $p\rightarrow c_1\cdots c_n$.

Figure~\ref{fOverview}b shows the overall architecture of TBCNN.
In our model, an AST node is first represented as a distributed,
real-valued vector so that the (anonymous) features of the symbols are captured.
The vector representations are learned by a coding criterion in our previous work
\cite{building}.

Then we design a set of subtree feature detectors, called the \textit{tree-based
convolution kernel}, sliding over the entire AST to extract structural information
of a program.
We thereafter apply dynamic pooling to gather information over different parts of the tree.
Finally, a hidden layer and an output layer are added. For supervised classification tasks,
the activation function of the output layer is $\softmax$.

In the rest of this section, we first explain the coding criterion for AST nodes'
representation learning, serving as a pretraining phase of programming language processing.
We then describe the proposed TBCNN model, including a coding layer, a convolutional
layer, and a pooling layer. We also provide additional information on dealing with nodes
that have varying numbers of child nodes, as in ASTs, by introducing the notion of continuous
binary trees.

 \subsection{Representation Learning for AST Nodes}\label{ssRepresentation}

Vector representations, sometimes known as \textit{embeddings}, can capture
underlying meanings of discrete symbols, like AST nodes. We propose in 
our previous work \cite{building} an unsupervised approach to learn program 
vector representations by a coding criterion, which serves as a way of pretraining.

A generic criterion for representation learning is ``smoothness''---similar symbols
have similar feature vectors \cite{RL}. For example, the symbols {\tt While} and
{\tt For} are similar because both of them are related to control flow, particularly
loops. But they are different from {\tt ID}, since {\tt ID} probably represents some data.
In our scenario, we would like the child nodes' representations to ``code'' their parent
node's via a single neural layer, during which both vector representations and
coding weights are learned. Formally, let $\vect(\cdot) \in \mathbb{R}^{N_f}$ be
the feature representation of a symbol, where $N_f$ is the feature dimension.
For each non-leaf node $p$ and its direct children $c_1, \cdots, c_n$, we would like
\begin{equation}
\vect(p) \approx \tanh\left(\sum\nolimits_i l_iW_{\text{code},i}\cdot \vect(c_i) + \bm b_{\text{code}}\right)
\label{eCode}
\end{equation}
\noindent where $W_{\text{code},i}\in\mathbb{R}^{N_f\times N_f}$ is the weight
matrix corresponding to the node $c_i$; $\bm b_{\text{code}}\in \mathbb{R}^{N_f}$ is the bias.
$l_i = \frac{\#\text{leaves under } c_i}{\#\text{leaves under } p}$ is the coefficient of the weight.
(Weights $W_{\text{code},i}$ are weighted by leaf numbers.)

Because different nodes may have different numbers of children, the number of
$W_{\text{code},i}$'s is not fixed. To overcome this problem, we introduce the
``continuous binary tree,'' where only two weight matrices $W_{\text{code}}^l$
and $W_{\text{code}}^r$ serve as model parameters. $W_i$ is a linear combination
of the two parameter matrices according to the position of node $i$. Details are deferred
to the last part of this section.

The closeness between $\vect(p)$ and its coded vector is measured by Euclidean distance square,
i.e.,
\begin{equation}\nonumber
 d = \left\|  \vect(p) - \tanh\left(\sum\nolimits_i l_iW_{\text{code}, i}\cdot \vect(c_i)+
 \bm b_{\text{code}}\right)\right\|_2^2
 \end{equation}

To prevent the pretraining algorithm from learning trivial representations (e.g., $\bm 0$'s
will give 0 distance but are meaningless), negative sampling is applied like \citeauthor{scratch} (\citeyear{scratch}).  For each pretraining data sample $p, c_1, \cdots, c_n$, we substitute
one symbol (either $p$ or one of $c$'s) with a random symbol. The distance of the negative sample
is denoted as $d_c$, which should be at least larger than that of the positive training sample
plus a margin $\Delta$ (set to 1 in our experiment). Thus, the pretraining objective is to
$$\operatorname*{minimize}\limits_{W_\text{code}^l,W_\text{code}^r,\bm b_{\text{code}},\vect(\cdot)}\ \max\left\{0, \Delta + d - d_c\right\}$$

\subsection{Coding Layer}\label{ssCoding}

Having pretrained the feature vectors for all symbols, we would like to feed them
forward to the tree-based convolutional layer for supervised learning. For leaf nodes,
they are just the vector representations learned in the pretraining phase. For a
non-leaf node $p$, it has two representations: the one learned in the pretraining
phase (left-hand side of Equation \ref{eCode}), and the coded one (right-hand side
of Equation \ref{eCode}). They are linearly combined before being fed to the convolutional
layer. Let $c_1, \cdots, c_n$ be the children of node $p$ and we denote the combined
vector as $\bm p$. We have
\begin{align}
\nonumber
\bm p &= W_{\text{comb1}}\cdot \vect(p)\\ \nonumber
 &+ W_{\text{comb2}}\cdot
\tanh\left(\sum\nolimits_{i} l_iW_{\text{code},i}\cdot\vect(x_i)+\bm b_{\text{code}}\right)
\end{align}
where $W_{\text{comb1}}, W_{\text{comb2}}\in\mathbb{R}^{N_f\times N_f}$ are the parameters
for combination. They are initialized as diagonal matrices and then fine-tuned during supervised training.

\subsection{Tree-based Convolutional Layer}

\begin{figure}[!t]
\centering
\includegraphics[width=2.5in]{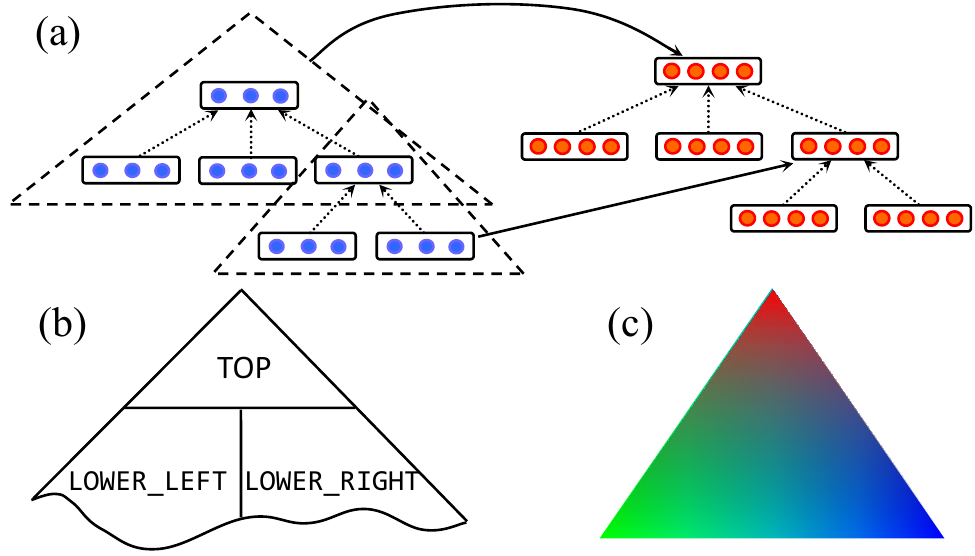}
\caption{(a) Tree-based convolution.
Nodes on the left are the feature vectors of AST nodes.
They are either pretrained or  combined with pretrained and coded vectors.
(b) An illustration of 3-way pooling.
(c) An analogy to the continuous binary tree model.
In the triangle, the color of a pixel is a combination of three primary colors;
in the convolution process, the weight for a node is a combination of three weight
parameters, namely $W_{\text{conv}}^t$, $W_{\text{conv}}^l$, and $W_{\text{conv}}^r$.}
\label{fConvolution}
\end{figure}

Now that each symbol in ASTs is represented as a distributed, real-valued
vector $\bm x \in \mathbb{R}^{N_f}$, we apply a set of fixed-depth feature
detectors sliding over the entire tree, depicted in Figure~\ref{fConvolution}a.
The subtree feature detectors can be viewed as convolution with a set of finite
support kernels. We call this \textit{tree-based convolution}.

Formally, in a fixed-depth window, if there are $n$ nodes with vector representations
$\bm x_1, \cdots, \bm x_n$, then the output of the feature detectors is\footnote{
We used {\tt tanh} as the activation function in TBCNN mainly because we hope to encode features to a same semantic space $(-1,1)$ during coding. We are grateful to an anonymous reviewer for reminding us of using {\tt ReLU} in convolution, and we are happy to try it in future work.}
\begin{equation}\nonumber
\bm y = \tanh\left(\sum\nolimits_{i=1}^n W_{\text{conv},i}\cdot\bm x_i +\bm b_{\text{conv}}\right)
\end{equation}
where $\bm y, \bm b_{\text{conv}}\!\!\in\!\mathbb{R}^{N_c}$, $W_{\text{conv},i}\!\!\in\!\mathbb{R}^{N_c\times N_f}$. ($N_c$ is the number of feature detectors.)
$\bm 0$'s are padded for nodes at the bottom that do not have as many layers as the feature detectors.
In our experiments, the kernel depth is set to 2.

Note that, to deal with varying numbers of children, we also adopt the notion of
continuous binary tree. In this scenario, three weight matrices serve as model parameters,
namely $W_{\text{conv}}^t$, $W_{\text{conv}}^l$, and $W_{\text{conv}}^r$.
$W_{\text{conv},i}$ is a linear combination of these three matrices (explained in detail
in the last part of this section).

\subsection{Dynamic Pooling}\label{ssPooling}

After convolution, structural features in an AST are extracted, and a new tree is generated.
The new tree has exactly the same shape and size as the original one, which is varying among
different programs. Therefore, the extracted features cannot be fed directly to a fixed-size neural layer.
Dynamic pooling \cite{dynamic} is applied to deal with this problem.

The simplest approach, perhaps, is to pool all features to one vector. We call this
\textit{one-way pooling}. Concretely, the maximum value in each dimension is taken from the
features that are detected by tree-based convolution. We also propose an alternative,
\textit{three-way pooling}, where features are pooled to 3 parts, TOP, LOWER\_LEFT,
and LOWER\_RIGHT, according to the their positions in the AST (Figure~\ref{fConvolution}b).
As we shall see from the experimental results, the simple one-way pooling
just works as well as three-way pooling. Therefore we adopt one-way pooling in our experiments.

After pooling, the features are fully connected to a hidden layer and then fed to the output layer
($\softmax$) for supervised classification. With the dynamic pooling process,
structural features along the entire AST reach the output layer with short paths. Hence,
they can be trained effectively by back-propagation.

\subsection{The ``Continuous Binary Tree'' Model}\label{ssCBT}

As stated, one problem of coding and convolving is that we cannot determine
the number of weight matrices because AST nodes have different numbers of children.

One possible solution is the continuous bag-of-words model
(CBoW, \citeauthor{word2vec}, \citeyear{word2vec}),\footnote{In their original paper, they do not deal with varying-length data, but
their method extends naturally to this scenario.
Their method is also mathematically equivalent to average pooling.
} but position information will be lost completely.
Such approach is also used in \citeauthor{multilingual} (\citeyear{multilingual}).
\citeauthor{grounded} (\citeyear{grounded})
allocate a different weight matrix as parameters for each position;
but this method fails to scale up since there will be a huge number of different positions in ASTs.

In our model, we view any subtree as a ``binary'' tree, regardless of its size and shape.
That is, we have only three weight matrices as parameters for convolution, and two for coding.
We call it a  \textit{continuous binary tree}.

Take convolution as an example. The three parameter matrices are
$W_{\text{conv}}^t$, $W_{\text{conv}}^l$,
and  $W_{\text{conv}}^r$. (Superscripts $t, l, r$ refer to ``top,''
``left,'' and ``right.'') For node $x_i$ in a window, its weight matrix
for convolution $W_{\text{conv},i}$ is a linear combination of
$W_{\text{conv}}^t$, $W_{\text{conv}}^l$, and  $W_{\text{conv}}^r$,
with coefficients $\eta_i^t$, $\eta_i^l$, and $\eta_i^r$, respectively.
The coefficients are computed according to the relative position of a node in the sliding window.
Figure~\ref{fConvolution}c is an analogy to the continuous binary tree model.
The equations for computing $\eta$'s are listed as follows.
\begin{compactitem}
\item $\eta_i^t=\frac{d_i-1}{d-1}$ ($d_i$: the depth of the node $i$ in the sliding window; $d$: the depth of the window.)
\item $\eta_i^r=(1-\eta_i^t)\frac{p_i-1}{n - 1}$. ($p_i$: the position of the node; $n$: the total number of $p$'s siblings.)
\item $\eta_i^l = (1-\eta_i^t)(1-\eta_i^r)$
\end{compactitem}

Likewise, the continuous binary tree for coding has two weight matrices
$W_{\text{code}}^l$ and $W_{\text{code}}^r$ as parameters. The details are not repeated here.

\medskip
To sum up, the entire parameter set for TBCNN is
$\Theta = \{W_{\text{code}}^l, W_{\text{code}}^r, W_{\text{comb1}},W_{\text{comb2}},
W_\text{conv}^t, W_\text{conv}^l,
W_\text{conv}^r, W_\text{hid},$ $
W_\text{out}, \bm b_{\text{code}}, \bm b_{\text{conv}}, \bm b_\text{hid}, \bm b_\text{out},
\vect(\cdot)\}$, where $W_{\text{hid}}$, $W_{\text{out}}$, $\bm b_\text{hid}$,
and $\bm b_\text{out}$
are the weights and biases for the hidden and output layers.
To set up supervised training, $W_{\text{code}}^l$, $W_{\text{code}}^r$,
$\bm b_{\text{code}}$, and $\vect(\cdot)$ are derived from the pretraining phase;
$W_{\text{comb1}}$ and  $W_{\text{comb2}}$ are initialized as diagonal matrices;
other parameters are initialized randomly. We apply the cross-entropy loss and use stochastic gradient descent, computed by back-propagation.

 \section{Experiments}
We first assess the learned vector representations both qualitatively and quantitatively. Then we evaluate TBCNN in two supervised learning tasks, and conduct model analysis.

The dataset of our experiments comes from a pedagogical programming open judge (OJ) system.\footnote{
http://programming.grids.cn
}
There are a large number of programming problems on the OJ system.
Students submit their source code as the solution to a certain problem;
the OJ system automatically judges the validity of submitted source code
by running the program.
We downloaded the source code and the corresponding programming problems (represented as IDs) as our dataset.

\subsection{Unsupervised Program Vector Representations}\label{ssPretrain}

We applied the coding criterion of pretraining to all {\tt C} code
in the OJ system, and obtained AST nodes' vector representations.

\begin{figure}[!t]\centering
\includegraphics[width=1.2in]{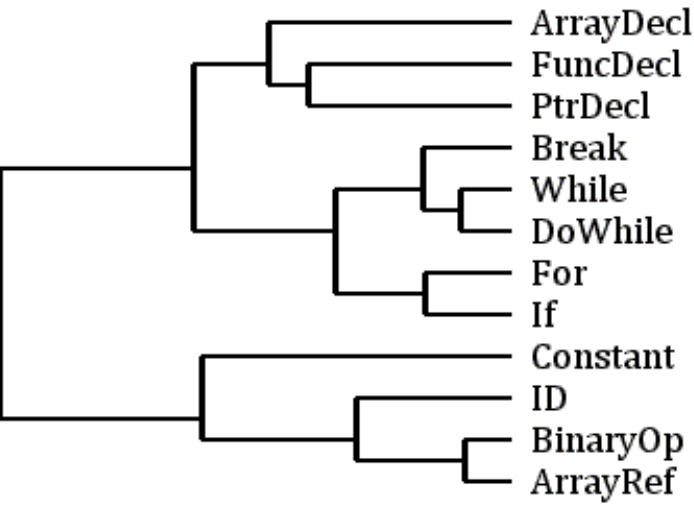}~(a)~~
\includegraphics[width=1.2in]{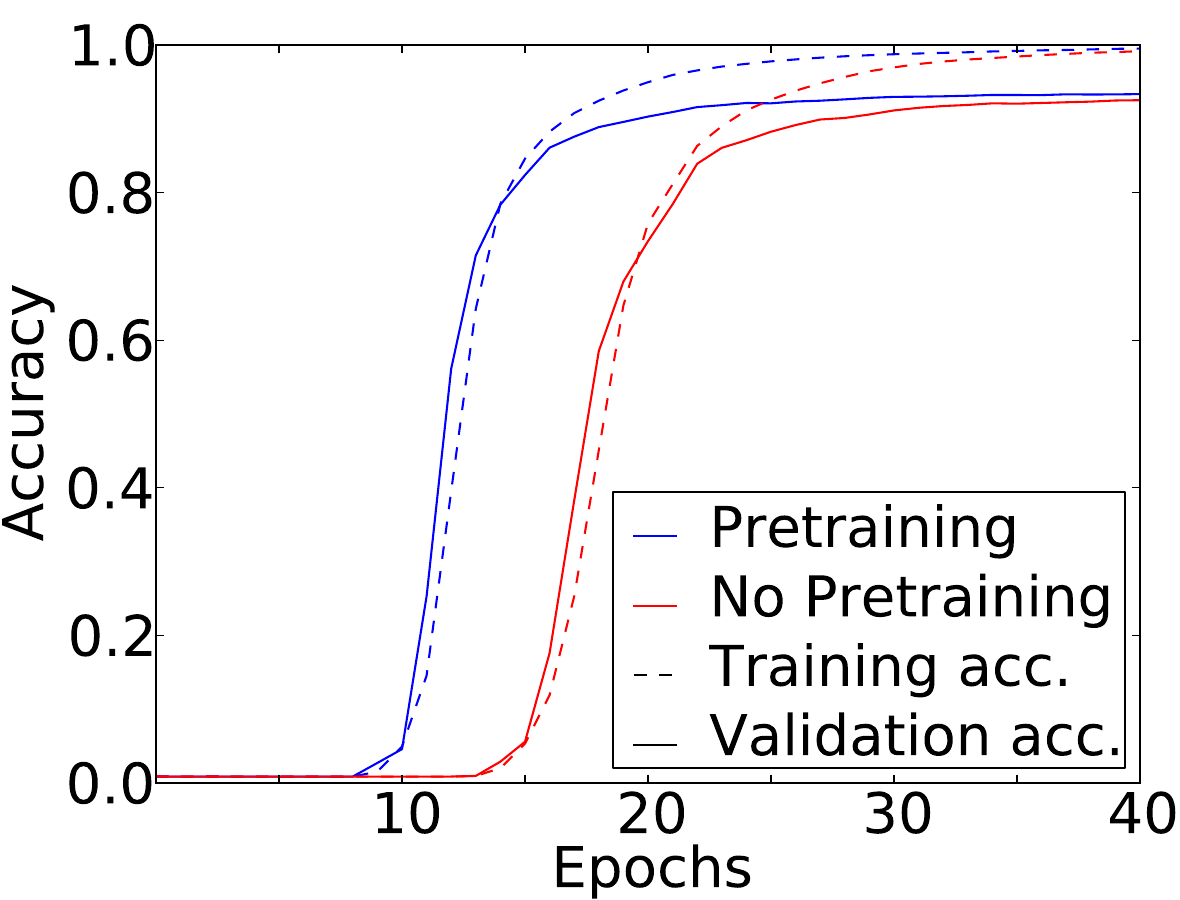}~~(b)
 \caption{Analysis of vector representations. (a) Hierarchical clustering based on AST nodes' vector representations. (b) Learning curves with and without pretraining.}
 \label{fHC}
 \end{figure}

\textbf{Qualitative analysis.}
Figure~\ref{fHC}a illustrates the hierarchical clustering result based on a subset of AST nodes.
As demonstrated, the symbols mainly fall into three categories:
(1) {\tt BinaryOp}, {\tt ArrayRef}, {\tt ID}, {\tt Constant} are grouped together since they are related
to data reference/manipulation; (2) {\tt For}, {\tt If},
{\tt While} are similar since they are
 related to control flow; (3) {\tt ArrayDecl}, {\tt FuncDecl},
 {\tt PtrDecl} are similar since they are declarations.
 The result is quite sensible because it is consistent with human understanding of programs.

 \textbf{Quantitative analysis.} We also evaluated pretraining's effect on supervised learning by feeding the learned representations to a program classification task. (See next subsection.)
 Figure~\ref{fHC}b plots the learning curves of both training and validation, which are compared with random initialization.
 Unsupervised vector representation learning
 accelerates the supervised training process by nearly $1/3$,
 showing that pretraining does capture underlying features
 of AST nodes, and that they can emerge high-level features spontaneously during supervised learning.
 However, pretraining has a limited effect on the final accuracy.
 One plausible explanation is that the number of AST nodes is small:
 the {\tt pycparser}, we use, distinguishes only 44 symbols.
 Hence, their representations can be adequately tuned in a supervised fashion.

 Nonetheless, we think the pretraining criterion
 is effective and beneficial for TBCNN,
 because training deep neural networks is usually time-consuming, especially
 when tuning hyperparameters. The pretrained vector representations
 are used throughout the experiments below.

% \begin{figure}[!t]
% \footnotesize
%~\!\includegraphics[width=.24\textwidth]{cluster7.pdf}~\hspace{-.23\textwidth}~(a)~\hspace{.2\textwidth}~\!\!\includegraphics[width=.235\textwidth]{pretraining.pdf}~\hspace{-.24\textwidth}~(b)
% \caption{Analysis of vector representations. (a) Hierarchical clustering based on AST nodes' vector representations. (b) Learning curves with and without pretraining.}
% \label{fHC}
% \end{figure}

 \begin{table}[!t]
  \centering
 \resizebox{.42\textwidth}{!}{
 \begin{tabular}{lrr}
 \hline
 \hline
 \textbf{Statistics} & \textbf{Mean} & \textbf{Sample std.}\\
 \hline
 $\#$ of code lines    &  36.3  & 19.0 \\
 $\#$ of AST nodes     &  189.6   & 106.0 \\
 Average leaf nodes' depth in an AST  &  7.6  &  1.7 \\
 Max depth of an AST   & 12.3  & 3.2\\
 \hline
 \hline
 \end{tabular}
 }
 \caption{Statistics of our dataset.}\label{statistics}
 \end{table}
 \begin{table}[!t]
 \centering
 \resizebox{.40\textwidth}{!}{
 \begin{tabular}{lll}
 \hline
 \hline
 \!\!\textbf{Hyperparameter} & \textbf{Value} & \textbf{How is the value chosen?}\!\!\\
 \hline
 \!\!Initial learning rate & 0.3 & By validation\!\!\\
 \!\!Learning rate decay   & None & Empirically\!\!\\
 \!\!Embedding dimension   & 30   & Empirically\!\!\\
 \!\!Convolutional layers' dim. & 600 & By validation\!\!\\
 \!\!Penultimate layer's dim. & 600   & Same as conv layers\!\!\\
 \!\!$l_2$ penalty  & None         & Empirically\!\!\\
 \hline
 \hline
 \end{tabular}
 }
  \caption{TBCNN's hyperparameters.}\label{figHyper}
 \end{table}
 \subsection{Classifying Programs by Functionalities}\label{ssClassify}

 \subsubsection{Task description}
 In software engineering, classifying programs by functionalities is an important problem
 for various software development tasks.
 For example, in a large software repository (e.g., SourceForge),
 software products are usually
 organized into categories, a typical criterion for which is by functionalities.
 With program classification, it becomes feasible to automatically tag a software component newly added into  the repository, which is beneficial for software reuse during the development process.

 In our experiment, we applied TBCNN to classify source code in the OJ system.
 The target label of a data sample is one of 104 programming problems (represented as an ID). That is, programs with a same target label have the same functionality.
 We randomly chose exactly 500 programs in each class, and thus 52,000 samples in total, which were further randomly split by 3:1:1 for training, validation,
 and testing. Relevant statistics are shown in Table \ref{statistics}.

\subsubsection{Hyperparameters}
TBCNN's hyperparameters are shown in Table \ref{figHyper}.
Our competing methods include SVM and a deep feed-forward neural network
based on hand-crafted features, namely bag-of-words (BoW, the counting of each
symbol) or bag-of-tree (BoT, the counting of 2-layer subtrees).
We also compare our model with the recursive neural network (RNN, \citeauthor{RAE} \citeyear{RAE}).
Hyperparameters for baselines are listed as follows.

\textbf{SVM.} The linear SVM has one hyperparameter $C$;
RBF SVM has two, $C$ and $\gamma$.
They are tuned by validation over the set $\{\cdots, 1, 0.3, 0.1, 0.03, \cdots\}$
with grid search.

\textbf{DNN.} We applied a 4-layer DNN (including input) empirically.
The hidden layers' dimension is 300, chosen from $\{100, 300, 1000\}$;
learning rates are 0.003 for BoW and 0.03 for BoT, chosen from $\{0.003, \cdots, 0.3\}$ with granularity $3$x.
$\ell_2$ regularization coefficient is $10^{-6}$ for both BoW and BoT,
chosen from $\{10^{-7}, \cdots, 10^{-4}\}$ with granularity $10$x, and also no regularization.

\textbf{RNN.} Recursive units are 600-dimensional, as in our method.
The learning rate is chosen from the set
$\{\cdots 1.0, 0.3, 0.1\cdots\}$, and 0.3 yields the highest validation performance.

  \begin{table}[!t]
  \centering
  \footnotesize
  \scalebox{.94}{  \begin{tabular}{c|c|c}
  \hline
  \hline
  \textbf{Group}& \textbf{Method} & \textbf{Test Accuracy (\%)}\\
  \hline
           & linear SVM$+$BoW         & 52.0 \\ %51.98\!\!\\
  Surface  & RBF SVM$+$BoW            & 83.9 \\ %83.89\!\!\\
  features & linear SVM$+$BoT         & 72.5 \\ %72.54\!\!\\
           & RBF SVM$+$BoT            & 88.2 \\ %88.30\!\!\\
  \hline
  			
             & DNN$+$BoW                  & 76.0\\ % 75.9615
  NN-based   & DNN$+$BoT                  & 89.7\\ % 89.6731
  approaches & Vector avg.                & 53.2\\ % 53.16
             & RNN                        & 84.8\\ % 84.80\!\!\\
  \hline
  Our method & TBCNN        &  \textbf{94.0}  \\ %\!\!\!\!\textbf{94.44}\!\!\!\!\\
  \hline
  \hline
  \end{tabular}}
  \caption{The accuracy  of 104-label program classifications.}
  \label{tClassification}
  \end{table}

\subsubsection{Results} Table \ref{tClassification} presents the results in the 104-label
program classification experiment.
Using SVM with surface features does distinguish different programs to some extent---for example,
a program about string manipulation is different from,
say, matrix operation; also, a difficult programming
problem necessitates a more complex program, and thus more lines of code and AST nodes.
However, their performance is comparatively low.

We tried deep feed-forward neural networks on these features, and achieved
accuracies of 76.0--89.7\%, comparable to SVMs. Vector averaging with $\softmax$---another neural network-based competing method applied in NLP \cite{RNN,CNNNLP}---yields an accuracy similar to a linear classifier built on BoW features. This is probably because the number of AST symbols is far fewer than words in natural languages, and thus the vector representations (provided non-singular)
can be absorbed into the classifier's weights.
Comparing these approaches with our method,
we deem TBCNN's performance boost is not merely caused by using a better classifier (neural networks versus SVM, say), but also the feature/representation learning nature, which enables automatic structural feature extraction.

We also applied RNN to the program classification task\footnote{We do not use the pretrained vector representations, which are
inimical to RNN: the weight $W_\text{code}$ codes children's representation to its
candidate parent's; adversely, the high-level nodes in programs (e.g., a function definition)
are typically non-informative.};
the RNN's accuracy is lower than shallow methods (SVM+BoT).
Taking into consideration experiments in NLP \cite{RAE,RNN},
we observe a degradation of RNN's performance
if the tree structure is large.

 TBCNN outperforms the above methods, yielding an accuracy of 94\%.
 By exploring tree-based convolution,
 our model is better at capturing programs' structural features, which
 is important for program analysis.

 \subsection{Detecting Bubble Sort}\label{ssBubble}

 \subsubsection{Task description} To further evaluate our TBCNN model in a more realistic SE scenario,
 we used it to detect an unhealthy code pattern, bubble sort, which
 can also be regarded as a (binary) program classification task.
 Detecting source code of certain patterns is closely related to many SE problems.
 In this experiment,
 bubble sort is thought of as unhealthy code because
 it implements an inefficient algorithm.
 By identifying such unhealthy code, project managers can refine the implementations
 during the maintenance process.

 Before the experiment, a volunteer\footnote{The volunteer
 has neither authorship nor a conflict of interests.} annotated, from the OJ system, 109 programs that contain bubble sort, and 109 programs that do not contain bubble sort. They were split 1:1 for validation and testing.

 \subsubsection{Data augmentation} To train our TBCNN model,
 a dataset of such scale is insufficient.
 We propose a simple yet useful data augmentation technique for programs.
 Concretely, we used the source code of 4k programs in the OJ system as the non-bubble sort class.
 For each program, we randomly substituted a fragment of program statements with a pre-written bubble sort snippet.
 Thus we had 8k data samples in total.

 %When testing the trained TBCNN in this experiment, we have two settings. In the first setting,
 %we used the mock data
 %for training, developing and testing, referred to as ``without transferring.''
 %In the second setting, we would like to see whether our model and data augmentation technique
 %can be applied to process real-world programs. We trained the model with the mock dataset, and
 %used the annotated dataset for developing and training (1:1 splitting).
 %Because the test set is written by real-world programmers, the styles and forms of bubble sort snippets may differ from the training set. We call this ``transfer learning.''

\begin{table}[!t]
\centering
 \begin{tabular}{llc}
 \hline
  \hline
  \textbf{Classifier} & \textbf{Features} & \textbf{Accuracy}\\
  \hline
  Rand/majority & -- & 50.0\\
  RBF SVM & Bag-of-words & 62.3\\ %62.30\\
  RBF SVM & Bag-of-trees & 77.1\\ %77.06\\
  TBCNN   & Learned     & \textbf{89.1}\\ % 89.09\\
  \hline
  \hline
  \end{tabular}
\caption{Accuracy of detecting bubble sort (in percentage).} \label{tBubble}
\end{table}
\begin{table}[!t]
\centering
 \scalebox{.90}{ \begin{tabular}{lc}
  \hline
  \hline
  \textbf{Model Variant} &\textbf{Validation Acc.}\\
  \hline
  Coding layer $\rightarrow$ None & 92.3\\
  1-way pooling $\rightarrow$ 3-way   & 94.3 \\
  Continuous binary tree $\rightarrow$ CBoW & 93.1 \\
  \hline
  TBCNN with the best gadgets & 94.4 \\
  \hline
  \hline
  \end{tabular}}
  \caption{Effect of coding, pooling, and the continuous binary tree.}
  \label{tVariant}
\end{table}
\begin{figure}[!t]
\centering
\includegraphics[width=2.3in]{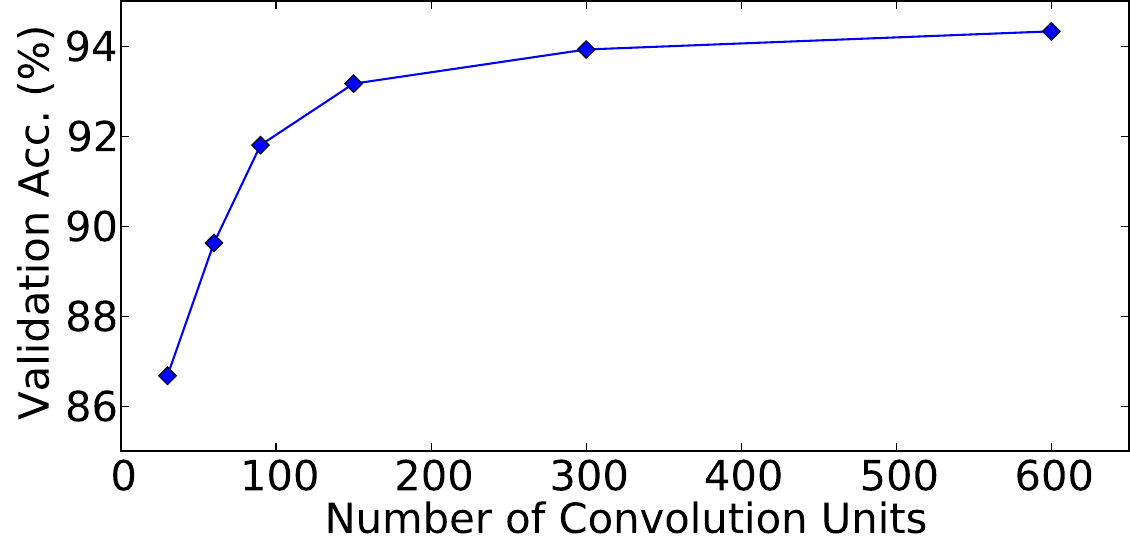}
\caption{Validation accuracy versus the number of convolution units.}\label{fConv}
\end{figure}

 \subsubsection{Results}
 %We first used the mock data for both training and testing.
 %TBCNN yields 99.68\% test accuracy, indicating that
 %our model can detect the pre-written bubble sort with very high
 %accuracy.
We tested our model on the annotated real-world programs. Note
that the test samples were written by real-world programmers, and thus
the styles and forms of bubble sort snippets may differ from the training set,
for example, sorting an integer array versus sorting a user-defined structure,
and sorting an array versus sorting two arrays simultaneously.
As we see in Table~\ref{tBubble}, bag-of-words features are not
illuminating in this classification
and yield a low accuracy of 62.3\%.
Bag-of-trees features are better, and achieve 77.06\%.
Our model outperforms these methods by more than 10\%.
This experiment also suggests that
neural networks can learn more robust features than
just counting surface statistics.

 \subsection{Model Analysis}\label{ssAnalysis}
 We now analyze each gadget of TBCNN quantitatively,
 with the 104-label program classification as our testbed.
 We report validation accuracies throughout this part.

 \subsubsection{Effect of coding layer} In the proposed
 TBCNN model for program analysis, we represent
 a non-leaf node by combining its coded representation and its pretrained one.
 We find that, the underneath coding layer can also integrate global information
 in addition to merely averaging two homogeneous sources.
 If we build a tree-based convolutional layer
 directly on the pretrained vector representations,
 all structural features are ``local,'' that is, confined in the convolution window.
 The lack of integrating global information leads to 2\%
 degradation in performance. (See the first and last rows in Table~\ref{tVariant}.)

 \subsubsection{Layers' dimensions}
 In our experiments, AST nodes' vector representations are set to be 30-dimensional empirically.
 We chose this small value because AST nodes have only 44 different symbols.
 Hence, the dimension needs to be, intuitively, smaller than words' vector
 representations, e.g., 300 in \citeauthor{tbcnn_sent} (\citeyear{tbcnn_sent}).
 The dimension of convolution, i.e., the number of feature detectors, was chosen
 by validation (Figure~\ref{fConv}). We tried several configurations, among which
 600-dimensional convolution results in the highest validation accuracy.
 This analysis also verifies that programs have rich structural information, even
 though the number of AST symbols is not large. As the rich semantics are emerged
 by different combinations of AST symbols, we are in need of
 more feature detectors, that is, a larger convolutional layer.

 \subsubsection{Effect of pooling layer}
 We tried two pooling methods in our TBCNN model,
 and compare them in Table~\ref{tVariant} (the second and last rows).
  3-way pooling  is proposed in hope of
 preserving features from different parts of the tree.
 However, as indicated by the experimental result,
 the simple 1-way pooling works just as fine (even 0.1\% higher on the validation set).
 This suggests that TBCNN is not sensitive to pooling methods,
 which mainly serve as a necessity for packing varying sized and shaped data.
 Further development can be addressed in future work.

 \subsubsection{Effect of continuous binary tree}
 The continuous binary tree is introduced to
 treat nodes with different numbers of children, as well as
 to capture order information of child nodes.
 We also implemented the continuous bag-of-words (CBoW) model, where
 child nodes' representations are averaged before convolution.
 Rows 4 and 5 in Table~\ref{tVariant} compare our proposed continuous binary tree
 and the above alternative.
 The result shows a boost of 1.3\% in considering
 child nodes' order information.

 \section{Conclusion}

 In this paper, we applied deep neural networks to the field of programming language processing.
 Due to the rich and explicit tree structures of programs,
 we proposed the novel Tree-Based Convolutional
 Neural Network (TBCNN).
 In our model, program vector representations are learned by the coding criterion;
 structural features are detected by the convolutional layer;
 the continuous binary tree and dynamic pooling enable our model to cope with trees of varying sizes and shapes.
Experimental results show the superiority of our model to baseline methods.
% The TBCNN model is evaluated in two program classification tasks; it outperforms
% baseline methods, including several NLP neural models.

\section{Acknowledgments}
We would like to thank anonymous reviewers for insightful comments; we also thank
Xiaowei Sun for annotating bubble sort programs, Yuxuan Liu for data processing, and Weiru Liu for discussion on the manuscript. This research is supported
by the National Basic Research Program of China
(the 973 Program) under Grant No. 2015CB352201 and the National Natural Science Foundation
of China under Grant Nos. 61421091, 61232015, 61225007, 91318301, and 61502014.

\fontsize{9.2pt}{10.2pt} \selectfont
%\small
\bibliographystyle{aaai}
\bibliography{dl,se}

\begin{thebibliography}{}

\bibitem[\protect\citeauthoryear{Bengio \bgroup et al\mbox.\egroup
  }{2006}]{laywise}
Bengio, Y.; Lamblin, P.; Popovici, D.; and Larochelle, H.
\newblock 2006.
\newblock Greedy layer-wise training of deep networks.
\newblock In {\em NIPS}.

\bibitem[\protect\citeauthoryear{Bengio, Courville, and Vincent}{2013}]{RL}
Bengio, Y.; Courville, A.; and Vincent, P.
\newblock 2013.
\newblock Representation learning: A review and new perspectives.
\newblock {\em IEEE Trans. Pattern Anal. Mach. Intell.} 35(8):1798--1828.

\bibitem[\protect\citeauthoryear{Bengio, Simard, and
  Frasconi}{1994}]{rnndifficult}
Bengio, Y.; Simard, P.; and Frasconi, P.
\newblock 1994.
\newblock Learning long-term dependencies with gradient descent is difficult.
\newblock {\em IEEE Trans. Neural Networks} 5(2):157--166.

\bibitem[\protect\citeauthoryear{Bettenburg and Begel}{2013}]{development}
Bettenburg, N., and Begel, A.
\newblock 2013.
\newblock Deciphering the story of software development through frequent
  pattern mining.
\newblock In {\em ICSE},  1197--1200.

\bibitem[\protect\citeauthoryear{Chilowicz, Duris, and Roussel}{2009}]{sim}
Chilowicz, M.; Duris, E.; and Roussel, G.
\newblock 2009.
\newblock Syntax tree fingerprinting for source code similarity detection.
\newblock In {\em Proc. IEEE Int. Conf. Program Comprehension},  243--247.

\bibitem[\protect\citeauthoryear{Collobert and Weston}{2008}]{unified}
Collobert, R., and Weston, J.
\newblock 2008.
\newblock A unified architecture for natural language processing: Deep neural
  networks with multitask learning.
\newblock In {\em ICML}.

\bibitem[\protect\citeauthoryear{Collobert \bgroup et al\mbox.\egroup
  }{2011}]{scratch}
Collobert, R.; Weston, J.; Bottou, L.; Karlen, M.; Kavukcuoglu, K.; and Kuksa,
  P.
\newblock 2011.
\newblock Natural language processing (almost) from scratch.
\newblock {\em JRML} 12:2493--2537.

\bibitem[\protect\citeauthoryear{Dahl, Mohamed, and Hinton}{2010}]{speech}
Dahl, G.; Mohamed, A.; and Hinton, G.
\newblock 2010.
\newblock Phone recognition with the mean-covariance restricted {Boltzmann}
  machine.
\newblock In {\em NIPS}.

\bibitem[\protect\citeauthoryear{Dietz \bgroup et al\mbox.\egroup }{2009}]{AI1}
Dietz, L.; Dallmeier, V.; Zeller, A.; and Scheffer, T.
\newblock 2009.
\newblock Localizing bugs in program executions with graphical models.
\newblock In {\em NIPS}.

\bibitem[\protect\citeauthoryear{Duvenaud \bgroup et al\mbox.\egroup
  }{2015}]{gbcnn}
Duvenaud, D.; Maclaurin, D.; Aguilera-Iparraguirre, J.; G{\'o}mez-Bombarelli,
  R.; Hirzel, T.; Aspuru-Guzik, A.; and Adams, R.
\newblock 2015.
\newblock Convolutional networks on graphs for learning molecular fingerprints.
\newblock {\em arXiv preprint arXiv:1509.09292}.

\bibitem[\protect\citeauthoryear{Ghabi and Egyed}{2012}]{requirement}
Ghabi, A., and Egyed, A.
\newblock 2012.
\newblock Code patterns for automatically validating requirements-to-code
  traces.
\newblock In {\em ASE},  200--209.

\bibitem[\protect\citeauthoryear{Hao \bgroup et al\mbox.\egroup
  }{2013}]{zhanglu}
Hao, D.; Lan, T.; Zhang, H.; Guo, C.; and Zhang, L.
\newblock 2013.
\newblock Is this a bug or an obsolete test?
\newblock In {\em Proc. ECOOP},  602--628.

\bibitem[\protect\citeauthoryear{Hermann and Blunsom}{2014}]{multilingual}
Hermann, K., and Blunsom, P.
\newblock 2014.
\newblock Multilingual models for compositional distributed semantics.
\newblock In {\em ACL},  58--68.

\bibitem[\protect\citeauthoryear{Hindle \bgroup et al\mbox.\egroup
  }{2012}]{naturalness}
Hindle, A.; Barr, E.; Su, Z.; Gabel, M.; and Devanbu, P.
\newblock 2012.
\newblock On the naturalness of software.
\newblock In {\em ICSE},  837--847.

\bibitem[\protect\citeauthoryear{Hinton, Osindero, and Teh}{2006}]{fast}
Hinton, G.; Osindero, S.; and Teh, Y.
\newblock 2006.
\newblock A fast learning algorithm for deep belief nets.
\newblock {\em Neural Computation} 18(7):1527--1554.

\bibitem[\protect\citeauthoryear{Kalchbrenner, Grefenstette, and
  Blunsom}{2014}]{CNNNLP}
Kalchbrenner, N.; Grefenstette, E.; and Blunsom, P.
\newblock 2014.
\newblock A convolutional neural network for modelling sentences.
\newblock In {\em ACL},  655--665.

\bibitem[\protect\citeauthoryear{Krizhevsky, Sutskever, and
  Hinton}{2012}]{imagenet}
Krizhevsky, A.; Sutskever, I.; and Hinton, G.
\newblock 2012.
\newblock {ImageNet} classification with deep convolutional neural networks.
\newblock In {\em NIPS}.

\bibitem[\protect\citeauthoryear{LeCun \bgroup et al\mbox.\egroup
  }{1995}]{lenet}
LeCun, Y.; Jackel, L.; Bottou, L.; Brunot, A.; Cortes, C.; Denker, J.; Drucker,
  H.; Guyon, I.; Muller, U.; and Sackinger, E.
\newblock 1995.
\newblock Comparison of learning algorithms for handwritten digit recognition.
\newblock In {\em Proc. Int. Conf. Artificial Neural Networks}.

\bibitem[\protect\citeauthoryear{Mikolov \bgroup et al\mbox.\egroup
  }{2013}]{word2vec}
Mikolov, T.; Sutskever, I.; Chen, K.; Corrado, G.; and Dean, J.
\newblock 2013.
\newblock Distributed representations of words and phrases and their
  compositionality.
\newblock In {\em NIPS}.

\bibitem[\protect\citeauthoryear{Mou \bgroup et al\mbox.\egroup
  }{2015}]{tbcnn_sent}
Mou, L.; Peng, H.; Li, G.; Xu, Y.; Zhang, L.; and Jin, Z.
\newblock 2015.
\newblock Discriminating neural sentence modeling by tree-based convolution.
\newblock In {\em EMNLP},  2315--2325.

\bibitem[\protect\citeauthoryear{Pane, Ratanamahatana, and Myers}{2001}]{PLNL}
Pane, J.; Ratanamahatana, C.; and Myers, B.
\newblock 2001.
\newblock Studying the language and structure in non-programmers' solutions to
  programming problems.
\newblock {\em Int. J. Human-Computer Studies} 54(2):237--264.

\bibitem[\protect\citeauthoryear{Peng \bgroup et al\mbox.\egroup
  }{2015}]{building}
Peng, H.; Mou, L.; Li, G.; Liu, Y.; Zhang, L.; and Jin, Z.
\newblock 2015.
\newblock Building program vector representations for deep learning.
\newblock In {\em Proc. 8th Int. Conf. Knowledge Science, Engineering and
  Management},  547--553.

\bibitem[\protect\citeauthoryear{Piech \bgroup et al\mbox.\egroup
  }{2015}]{hoare}
Piech, C.; Huang, J.; Nguyen, A.; Phulsuksombati, M.; Sahami, M.; and Guibas,
  L.
\newblock 2015.
\newblock Learning program embeddings to propagate feedback on student code.
\newblock In {\em ICML}.

\bibitem[\protect\citeauthoryear{Pinker}{1994}]{instinct}
Pinker, S.
\newblock 1994.
\newblock {\em The Language Instinct: The New Science of Language and Mind}.
\newblock Pengiun Press.

\bibitem[\protect\citeauthoryear{Socher \bgroup et al\mbox.\egroup
  }{2011a}]{dynamic}
Socher, R.; Huang, E.; Pennin, J.; Manning, C.; and Ng, A.
\newblock 2011a.
\newblock Dynamic pooling and unfolding recursive autoencoders for paraphrase
  detection.
\newblock In {\em NIPS}.

\bibitem[\protect\citeauthoryear{Socher \bgroup et al\mbox.\egroup
  }{2011b}]{RAE}
Socher, R.; Pennington, J.; Huang, E.; Ng, A.; and Manning, C.
\newblock 2011b.
\newblock Semi-supervised recursive autoencoders for predicting sentiment
  distributions.
\newblock In {\em EMNLP},  151--161.

\bibitem[\protect\citeauthoryear{Socher \bgroup et al\mbox.\egroup
  }{2013}]{RNN}
Socher, R.; Perelygin, A.; Wu, J.; Chuang, J.; Manning, C.; Ng, A.; and Potts,
  C.
\newblock 2013.
\newblock Recursive deep models for semantic compositionality over a sentiment
  treebank.
\newblock In {\em EMNLP},  1631--1642.

\bibitem[\protect\citeauthoryear{Socher \bgroup et al\mbox.\egroup
  }{2014}]{grounded}
Socher, R.; Karpathy, A.; Le, Q.; Manning, C.; and Ng, A.~Y.
\newblock 2014.
\newblock Grounded compositional semantics for finding and describing images
  with sentences.
\newblock {\em TACL} 2:207--218.

\bibitem[\protect\citeauthoryear{Steidl and Gode}{2013}]{bug}
Steidl, D., and Gode, N.
\newblock 2013.
\newblock Feature-based detection of bugs in clones.
\newblock In {\em 7th Int. Workshop on Software Clones},  76--82.

\bibitem[\protect\citeauthoryear{Zaremba and Sutskever}{2014}]{execute}
Zaremba, W., and Sutskever, I.
\newblock 2014.
\newblock Learning to execute.
\newblock {\em arXiv preprint arXiv:1410.4615}.

\end{thebibliography}

\end{document}